# A generic rule-based system for clinical trial patient selection

**Jianlin Shi[1], Kevin Graves[1], John F. Hurdle[1]**
**1 Department of Biomedical Informatics, University of Utah, Salt Lake City, US**

## ABSTRACT

*The n2c2 2018 Challenge task 1 aimed to identify patients who meet lists of heterogeneous inclusion/exclusion criteria for a hypothetical clinical trial. We demonstrate a generic rule-based natural language pipeline can support this task with decent performance (the average F1 score on the test set is 0.89, ranked the 9th out of 45 teams ).*

## INTRODUCTION

Recruiting enough patients is the precondition for a clinical trial to success. Around 19% of clinical trials failed because of insufficient participants.[1] One of the most efficient recruiting approaches is through using medical records. [2] However, because of the heterogeneity of eligibility criteria, a large amount of the health record information related to the eligibility criteria does not exist in a structured format, such as signs and symptoms.[3] On the other side, clinical notes often record additional information, including a patient's information prior to a visit. For instance, lab results are usually stored in a structured format during a visit but absent if the labs are done outside the health provider's network. Natural language processing (NLP) can extract this information to improve the efficiency of patient selection significantly.[4]

Utilizing health information in clinical notes have been demonstrated to be effective for clinical trial patient selection in previous studies.[4–7] Compared with general patient cohort identification,[8,9] clinical trial patient selection is relatively more challenging when eligible criteria are complex and often results in lower precision, even the underline technologies are largely overlapped. [4,5] Regarding the text information extraction, there are three types of classical approaches: rule-based, machine learning based (ML-based), and hybrid. Each of them has its pros and cons. Given recent advances of deep learning in NLP, some of these deep learning techniques have been experienced in clinical fields.[10] However, like many other ML-based approaches, the interpretability is still a significant limitation.[10] Under the clinical trial patient selection scenario, the interpretability is not as critical as applications in clinical decision support but still beneficial. With interpretable system outputs, researchers can quickly validate a system selected patient by reviewing system outputs without reading all his records. Additionally, developing the annotated data for ML-based model to train is still labor intensive. On the contrary, the rule-based approach often results in intensive labor for rule developing and slow speed when rules become large in amount.[11] Moreover, engineering rules becomes challenging when complex logic involves.[11] Thus, hybrid approaches are often used in these cases.

In this paper, we present our purely rule-based solution for clinical trial patient selection in the 2018 National NLP Clinical Challenges (n2c2) track 1. Keeping the inherent transparency of the rule-based approach, we enhanced the interpretability through a step-by-step visualization for each NLP component's outputs. We used an optimized rule processing engine to overcome the execution speed shortcoming[12] and heuristically designed inference steps to reduce the rule development complexity. Additionally, we used a semi-automated dictionary builder to reduce the rule-development labor. This solution has been applied in other patient cohort identification,[13,14] which demonstrated its generalizability.

## METHOD

### Dataset

The dataset was created and provided by the N2C2 challenge organizer, including 288 patients with 2-5 longitudinal clinical records. The training set contained 202 patients, and the test set contained 86 patients. These documents consisted of various types, such as History and Physical (H&P), Discharge Summary, and Emails. Each patient's records were concatenated into a single file with annotations in XML format.

The aim The Challenge task aims to identify patients who meet lists of inclusion/exclusion criteria for a hypothetical clinical trial. A wide range of eligibility criteria (13 in total) were specified, including demographic information, social history, mental status, comorbidities, medication usage, and lab values.

### Preprocessing

In order to make our solution more generalizable and close to real-world settings, we converted the data into EMR-like structure: 1) split the records from each file, 2) generated record IDs and patient IDs to keep track of the ownership, 3) imported the records into a database table, 4) extract the record date from each record and imported to a date column for each record inside the database, 5) infer the reference date as the latest record date (per annotation guideline), and 6) imported to a reference date column.

### Rule-based NLP Architecture

We enhanced our previous EasyCIE[13] NLP components with section detector and patient inferencer (see Figure 1). The specific function for each component was explained the Table 1. EasyCIE is a user-friendly rule-based clinical NLP tool. Its backend NLP components are using an n-trie based rule processing engine [12], which significantly improves the rule execution speed. EasyCIE also allows step-by-step NLP component debugging to support rule development (see Figure 2).

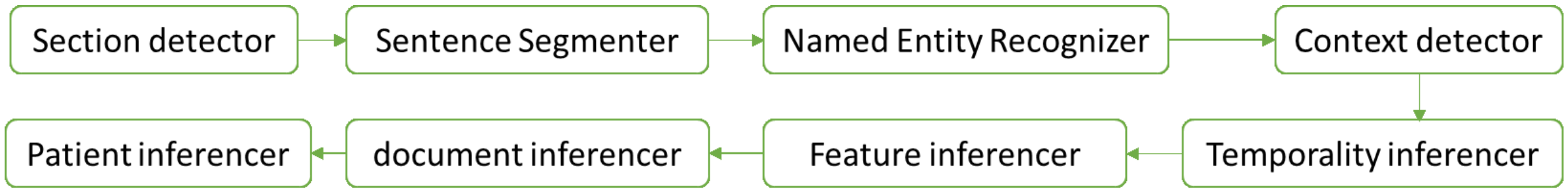

**Figure 1.** The components and workflow of the rule-based pipeline.

**Simplified rule syntaxes**

To facilitate rapid rule development, we redesigned the rule representations using simplified grammars and syntaxes instead of using any existing rule syntax, such as Drools. The core idea is to remove as much syntax as possible, such as "if," "then," by placing the actual rule elements in predefined cells in a row of csv or excel file. For example, in Figure 3, the vocabulary was placed in the first column, the conclusion type was in the second column, and the attribute regarding inclusion or

**Table 1.** The functions of NLP components

| **Component name** | **Functions** |
|---|---|
| Section Detector | To detect sections of interest* |
| Sentence Segmenter | A Rule-based sentence Segmenter using Hashing (RuSH)* |
| Named Entity Recognizer | To match a customizable dictionary with the support of exclusion dictionaries* |
| Context Detector | An optimized implementation of the context algorithm, attach the identified context value to corresponding named entities* |
| Temporality Inferencer | To recognize temporal statements (e.g., in the early 90s) within the context of named entities and compares the identified date against the reference date (e.g., admission date) to make the decision whether the corresponding named entities are historical mentions or presents mentions* |
| Feature Inferencer | To draw a local conclusion of a named entity, keep or assign a new type name based on its context values* |
| Document Inferencer | To draw a document level conclusion based on the all the mention level annotations in a document |
| Patient Inferencer | To draw a patient level conclusion based on the document level conclusions from multiple documents. |

* These components were implemented using n-trie (NLP-trie) [12] to optimize the execution speed.

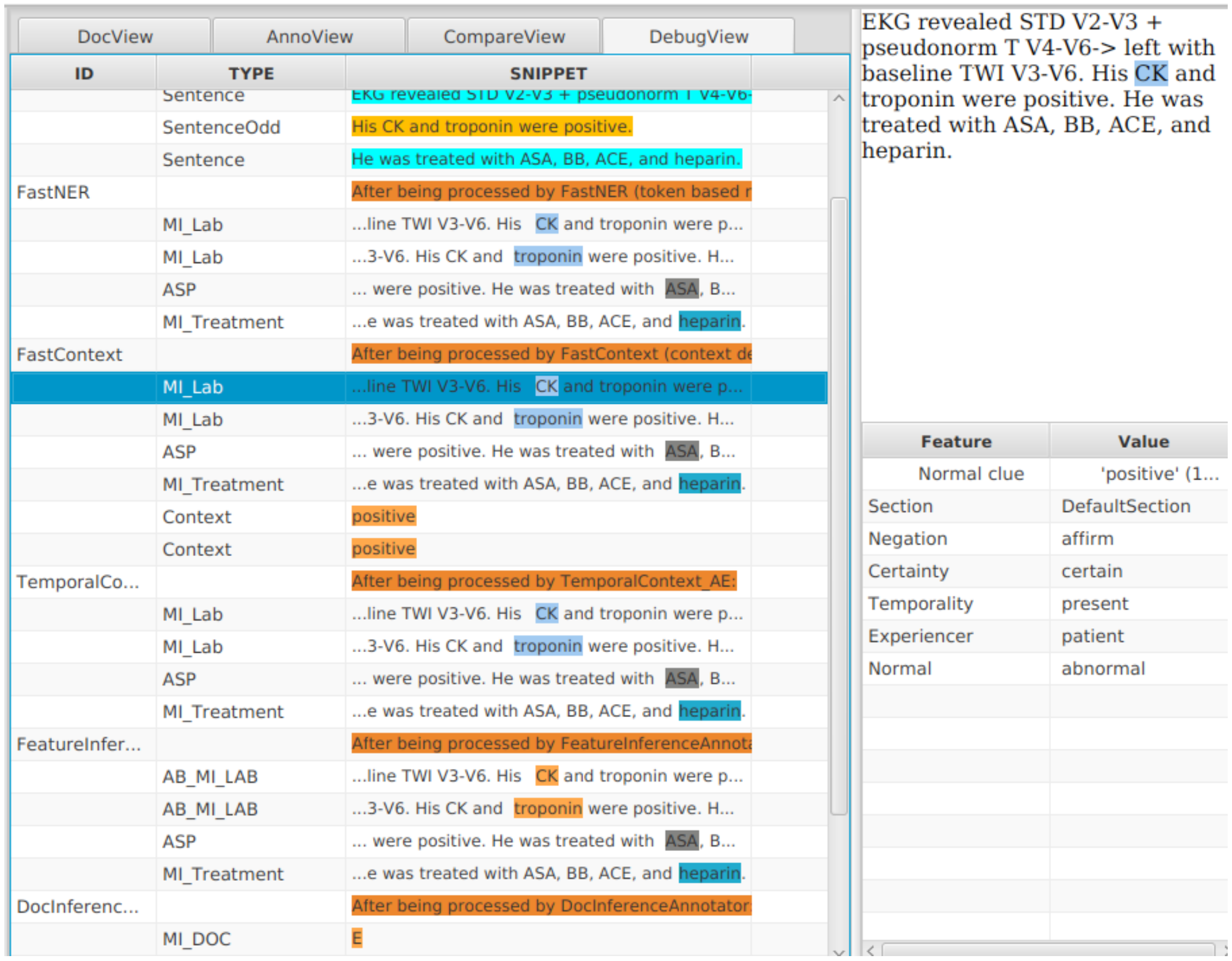

**Figure 2.** A screenshot of EasyCIE's step-by-step debugging function for NLP rule development.

exclusion was in the third one. Thus, the second row can represent the rule: if the phrase "myocardial infarct" is found in a document, annotate the phrase as "MI." In this way, a rule can be expressed more concisely. Additionally, by using copy and paste functions in Excel, users can easily create a large number of similar rules.

Using a similar format, we also simplified the inference rules. For example, in the rules of the Feature Inferencer (**Figure** 4), the logic conjunctions were expressed through list all logic elements joined by commas, while logic disjunctions were expressed by list all the elements across different rows. Specifically, in the second row, the rule meant if an annotation

| | A | B | C |
|---|---|---|---|
| 1 | **Vocab** | **Concept** | **Type** |
| 2 | myocardial infarct | MI | ACTUAL |
| 3 | infarction | MI_Candidate | ACTUAL |
| 4 | Mi / day | MI | PSEUDO |

**Figure 3.** A screen shot of rules for the name entity recognizer. The first rule means if a phrase "myocardial infarct" is found, label the phrase as "MI" concept. The last row means if a phrase "Mi/day" is found, do not label it as "MI" concept. By using this predefined structure, this rule syntax can avoid typing assistant words, e.g. "if" and "then", to make the rule clearer in appearance and easier to edit.

"MI_Candidate" (myocardial infarction candidate) in "Findings" section has attribute values: "affirm," "certain," "patient" and "cardiac," then this annotation is an "MI." The conclusion will copy all the corresponding attribute values. If put three rows together, the three rules meant if a "MI_Candidate" with these attribute values located in "Findings," "Impression," or "PresentHistory" section, this "MI_Candidate" is an "MI."

| | A | B | C | D | E |
|---|---|---|---|---|---|
| 1 | **Conclusion** | **ConclusionAttributes** | **Evidence** | **EvidenceAttributes** | **Section** |
| 2 | MI | COPYALL | MI_Candidate | affirm,certain,patient,cardiac | Findings |
| 3 | MI | COPYALL | MI_Candidate | affirm,certain,patient,cardiac | Impression |
| 4 | MI | COPYALL | MI_Candidate | affirm,certain,patient,cardiac | PresentHistory |

**Figure 4.** A screenshot of rules for the feature inferencer. For instance, the first rule means if a "MI_Canidate" concept with attributes values: "affirm," "certain," "patient," and "cardiac," and this concept is found is "Findings" section, then label this concept as "MI" concept.

We wrapped this pipeline into our EasyCIE to facilitate rapid rule development. For NER rules, we semi-automated dictionary building using UMLS and clinical word embeddings to find synonyms and closely related words. Rules were initially based on the annotation guideline and then improved by studying the training dataset.

## RESULTS AND DISCUSSION

Our test set micro-average F score achieved 0.884 (detailed scores are listed in Table 2). Because of the transparency of the rule-based approach and debugging assistance provided by EasyCIE, we are able to compare the NLP output against the gold standard easily. Some errors were introduced by the dictionary that we imported from online resources. For example, "Taiwanese" is not included as a language name. Some other errors were caused by the unexpected context of symptoms mentions (e.g., "apply to **thick skin** on feet"). Due to the absence of the mention-level gold standard, we are not able to investigate all the false negatives. However, we still identified a fair number of *arguable* false positive instances (13 patients). Some of them were confirmed by the organizers. For instance, a patient has CAD history, was treated by "nitroglycerin" in one note, and "labetalol" in another note. The original annotation was "not met," while the true answer should be "met" according to annotation guideline.

## CONCLUSION

Our single rule-based pipeline provides a decent solution for this NLP challenge task with heterogeneous nature, demonstrating its flexibility and genericity to be applicable for patient cohort identification.

**Table 2.** The performance of each criteria and overall task

| | met | | | | not met | | | overall | |
|---|---|---|---|---|---|---|---|---|---|
| | Prec. | Rec. | Speci. | F(b=1) | Prec. | Rec. | F(b=1) | F(b=1) | AUC |
| Abdominal | 0.8621 | 0.8333 | 0.9286 | 0.8475 | 0.9123 | 0.9286 | 0.9204 | 0.8839 | 0.8810 |
| Advanced-cad | 0.8400 | 0.9333 | 0.8049 | 0.8842 | 0.9167 | 0.8049 | 0.8571 | 0.8707 | 0.8691 |
| Alcohol-abuse | 0.0000 | 0.0000 | 0.9639 | 0.0000 | 0.9639 | 0.9639 | 0.9639 | 0.4819 | 0.4819 |
| Asp-for-mi | 0.8354 | 0.9706 | 0.2778 | 0.8980 | 0.7143 | 0.2778 | 0.4000 | 0.6490 | 0.6242 |
| Creatinine | 0.6923 | 0.7500 | 0.8710 | 0.7200 | 0.9000 | 0.8710 | 0.8852 | 0.8026 | 0.8105 |
| Dietsupp-2mos | 0.8039 | 0.9318 | 0.7619 | 0.8632 | 0.9143 | 0.7619 | 0.8312 | 0.8472 | 0.8469 |
| Drug-abuse | 0.2500 | 0.6667 | 0.9277 | 0.3636 | 0.9872 | 0.9277 | 0.9565 | 0.6601 | 0.7972 |
| English | 0.9211 | 0.9589 | 0.5385 | 0.9396 | 0.7000 | 0.5385 | 0.6087 | 0.7741 | 0.7487 |
| Hba1c | 0.9630 | 0.7429 | 0.9804 | 0.8387 | 0.8475 | 0.9804 | 0.9091 | 0.8739 | 0.8616 |
| Keto-1yr | 0.0000 | 0.0000 | 1.0000 | 0.0000 | 1.0000 | 1.0000 | 1.0000 | 0.5000 | 0.5000 |
| Major-diabetes | 0.7273 | 0.7442 | 0.7209 | 0.7356 | 0.7381 | 0.7209 | 0.7294 | 0.7325 | 0.7326 |
| Makes-decisions | 0.9765 | 1.0000 | 0.3333 | 0.9881 | 1.0000 | 0.3333 | 0.5000 | 0.7440 | 0.6667 |
| Mi-6mos | 0.5000 | 0.6250 | 0.9359 | 0.5556 | 0.9605 | 0.9359 | 0.9481 | 0.7518 | 0.7804 |
| Overall (micro) | 0.8402 | 0.8932 | 0.8816 | 0.8659 | 0.9222 | 0.8816 | 0.9015 | 0.8837 | 0.8874 |
| Overall (macro) | 0.6440 | 0.7044 | 0.7727 | 0.6642 | 0.8888 | 0.7727 | 0.8084 | 0.7363 | 0.7385 |